\documentclass[conference]{IEEEtran}
\IEEEoverridecommandlockouts
\usepackage{cite}
\usepackage{amsmath,amssymb,amsfonts}
\usepackage{graphicx}
\usepackage{textcomp}
\usepackage[dvipsnames]{xcolor} 
\usepackage{booktabs}
\usepackage{multirow}
\usepackage{microtype}
\usepackage{hyperref}
\usepackage{cleveref}
\usepackage{algorithm}
\usepackage{caption} 
\usepackage{subcaption} 
\usepackage{balance}
\usepackage[colorinlistoftodos,prependcaption,disable]{todonotes} 
\setuptodonotes{inline}
\usepackage{algorithmic}

\usepackage{tikz}

\IEEEoverridecommandlockouts
\IEEEpubid{\makebox[\columnwidth]{ 979-8-3503-5067-8/24/\$31.00~\copyright2024 IEEE \hfill} 
\hspace{\columnsep}\makebox[\columnwidth]{ }}

\def\BibTeX{{\rm B\kern-.05em{\sc i\kern-.025em b}\kern-.08em
    T\kern-.1667em\lower.7ex\hbox{E}\kern-.125emX}}
\begin{document}

\title{Online Optimization of Curriculum Learning Schedules using Evolutionary Optimization}



\author{\IEEEauthorblockN{Mohit Jiwatode, Leon Schlecht, Alexander Dockhorn}
\IEEEauthorblockA{Institute for Information Processing\\
Leibniz University Hannover\\
Hannover, Germany\\
\{jiwatode, schlecht, dockhorn\}@tnt.uni-hannover.de}
}

\maketitle
\IEEEpeerreviewmaketitle

\IEEEpubidadjcol

\todo{CHECK for consistency in the way Minigrid DoorKey DynamicObstacles RHEA CL/RHEA-CL is written}

\begin{abstract}
We propose RHEA CL, which combines Curriculum Learning (CL) with Rolling Horizon Evolutionary Algorithms (RHEA) to automatically produce effective curricula during the training of a reinforcement learning agent. RHEA CL optimizes a population of curricula, using an evolutionary algorithm, and selects the best-performing curriculum as the starting point for the next training epoch. Performance evaluations are conducted after every curriculum step in all environments. We evaluate the algorithm on the \textit{DoorKey} and \textit{DynamicObstacles} environments within the Minigrid framework. It demonstrates adaptability and consistent improvement, particularly in the early stages, while reaching a stable performance later that is capable of outperforming other curriculum learners. In comparison to other curriculum schedules, RHEA CL has shown to yield performance improvements for the final Reinforcement learning (RL) agent at the cost of additional evaluation during training.
\todo{Add some results here}
\end{abstract}


\section{Introduction}
\label{sec:introduction}

\todo[inline]{we need to add a bit more information about why this is relevant for Game AI, e.g. general performance across multiple levels, increased learning speed, better final performance}

\todo{can the hyperparameter landscapes be added here}


Machine Learning (ML) algorithms often get trapped in local minima during optimization, making obtaining optimal solutions challenging. CL addresses this by presenting training data in a meaningful order, allowing agents to learn progressively, and emulating human-like learning. This method has shown improvements in generalizability and convergence across various domains such as computer vision and reinforcement learning.


CL has proven effective in various fields including computer vision, natural language processing, and reinforcement learning~\cite{soviany2022curriculum} \cite{wang2021survey}. Traditional curriculum design, often based on simple rules, may not effectively address the complexity of tasks or adapt to evolving model needs, leading to poorer outcomes. Automated methods could enhance curriculum scheduling.


In continuous decision-making, RHEA have successfully guided agent actions within limited planning horizons, performing well across diverse tasks and matching other top solutions~\cite{gaina2021rolling}. We introduce a new RHEA-based curriculum scheduler to dynamically optimize curricula, aiming to circumvent training pitfalls and boost final performance through continuous curriculum evaluation.

In summary, our main contributions are:
\begin{itemize}    
    \item We propose the RHEA CL algorithm, a curriculum optimization algorithm that selects the locally optimal environment to continue training.
    \item The presented algorithm is evaluated and compared with other curriculum learning algorithms based on two environments of the Minigrid framework.
    \item Furthermore, we analyze the algorithm's parameter space to get insights into its optimal configuration.
\end{itemize}

Integrating RHEA with CL presents a significant advancement in the development of Game AI. This combination leverages the adaptive and predictive capabilities of RHEA to dynamically adjust the curriculum, ensuring that game agents are not only exposed to progressively challenging scenarios but also that these scenarios are optimized for the agent's current learning stage. The RHEA CL approach allows for a more personalized learning trajectory, tailoring the difficulty and content of the learning materials to the agent's evolving abilities. This method has the potential to significantly enhance the learning speed, and general performance across various game levels, and achieve better final performance in complex game environments. 

\todo{Results are removed from introduction}

Our research further examines the diverse impacts that the hyperparameters of Evolutionary Algorithms (EAs), Rolling Horizons (RH), and CL have on overall performance and rates of convergence. We compare the performance of our algorithm with state-of-the-art baselines and report the results. Furthermore, we discuss the limitations of our work and propose interesting directions for future research.


\begin{figure*}[t]
     \centering
     \begin{subfigure}[c]{0.15\textwidth}
         \centering
         \includegraphics[width=0.9\textwidth]{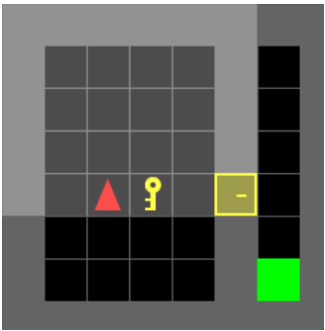}
         \caption{DoorKey 8x8}
         \label{fig:DoorKey 8x8}
     \end{subfigure}
     \begin{subfigure}[c]{0.23\textwidth}
         \centering
         \includegraphics[width=0.9\textwidth]{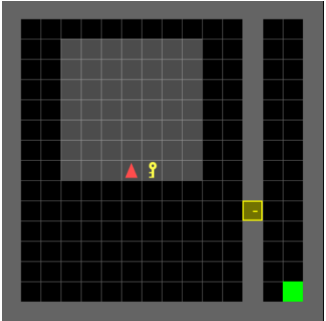}
         \caption{DoorKey 16x16}
         \label{fig:DoorKey 16x16}
     \end{subfigure}
     \hfill
      \begin{subfigure}[c]{0.23\textwidth}
         \centering
         \includegraphics[width=0.6\textwidth]{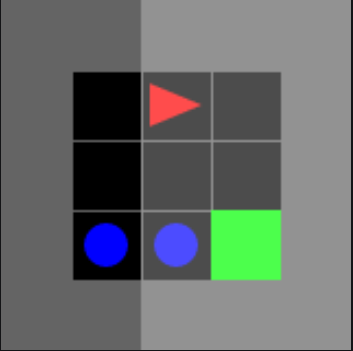}
         \caption{DynamicObstacles 5x5}
         \label{fig:DynamicObsctacle 5x5}
     \end{subfigure}
          \begin{subfigure}[c]{0.22\textwidth}
         \centering
         \includegraphics[width=0.9\textwidth]{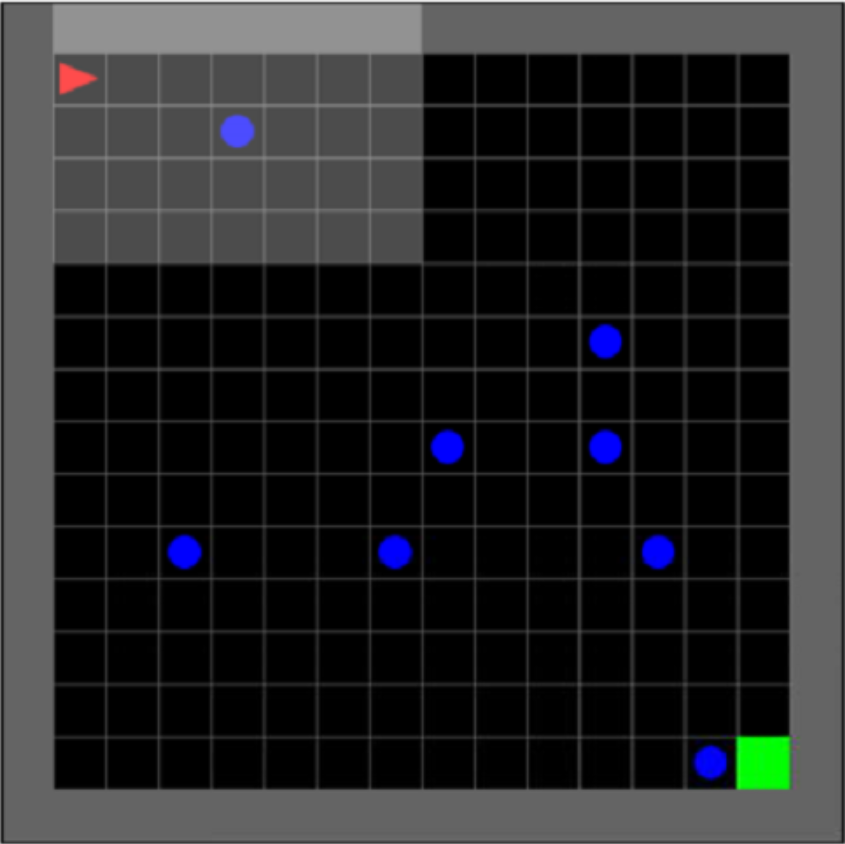}
         \caption{DynamicObstacles 16x16}
         \label{fig:dynamic obstacle 16x16}
     \end{subfigure}
        \caption{Different sizes of the DoorKey and DynamicObstacles environments. The lighter area indicates the observation space of the agent.}
        \label{fig:Different sizes of the Minigrid}
\end{figure*}

\section{Background}

RHEA CL uses CL, EAs, and RH to optimize the learning process of a RL agent. These concepts serve as the backbone of our work and will be briefly introduced in this section.

\subsection{Curriculum Learning}

Bengio et al.~\cite{bengio2009curriculum} defined CL as a training strategy for ML algorithms using a curriculum of tasks. More formally \cite{wang2021survey}, CL is a training criteria sequence over training steps $T$: $C = \,  \langle Q_1,..., Q_t,...Q_T \rangle $ where every $Q_t$ is (re)weighed target training distribution ($P(z)$) over examples $z$ of the training set $D$ at time $t$:
\begin{equation}
    Q_t(z) \propto W_t(z)P(z)  \\: \forall z \in D.
\end{equation}
Such that:
\begin{enumerate}
    \item The distribution entropy ($H$) increases with time $t$ i.e., $H(Q_t) < H(Q_{t+1})$.
    \item Sample weight ($W_t$) increases with time $t$ i.e.,\newline $W_t \le W_{t+1}$.
    \item $Q_T(z) = P(z)$ 
\end{enumerate} 
The two main tasks of CL are (i) ranking samples based on their complexity from easy to hard, and (ii) determining the optimal pacing for introducing these samples to the learning process \cite{soviany2022curriculum}.

\subsection{Rolling Horizon Evolutionary Algorithm}

Rolling horizon decision-making~\cite{sethi1991theory} is a strategy used in dynamic, uncertain environments, which require frequent evaluation and optimization of planned action sequences. For each decision, the agent creates a plan up to a defined forecast horizon. Once the plan has been optimized only its first action is applied to the environment and the remaining sequence is used to initialize the search for the next plan. Thus, the horizon effectively rolls forward with each executed action.

The RHEA algorithm~\cite{gaina2021rolling} uses evolutionary optimization~\cite{holland1992adaptation} to find the best action sequence at any given time step. For each queried action, a population of action sequences is generated and modified across multiple generations. Once the evolutionary optimization terminates, the first action of the current population's best action sequence is applied. The remaining action sequence is once again used to initialize the individuals of the next population.

\subsection{Reinforcement Learning}

RL ~\cite{sutton2018reinforcement} is a paradigm of machine learning in which an autonomous agent is tasked with finding a sequence of actions in an environment to optimize cumulative reward during an episode. The agent undergoes many iterations of trial and error to find a mapping from states to optimal actions, called policy.

The Proximal Policy Optimization (PPO) algorithm~\cite{schulman2017proximal} works by iteratively collecting data from the environment using the current policy, calculating advantages to estimate the quality of each action compared to the current policy. Then it updates the policy parameters using a surrogate objective that encourages rather small policy updates to prevent large policy changes, which could worsen performance. The algorithm optimizes for a balance between exploring new actions with the help of an entropy regularization term and maximizes the expected cumulative reward. Simultaneously, it ensures that the policy updates do not deviate too far from the previous policy (clipping the policy ratio). The combination of exploration and stability makes PPO an effective algorithm for a wide range of RL tasks.

\subsection{Minigrid Environments}

The Minigrid environment~\cite{chevalier2023minigrid} maintained by the Farama Foundation, offers a range of grid-based worlds for agents to perform tasks such as object interaction, navigation, and obstacle avoidance. The environments primarily use sparse rewards that are given when agents achieve their goals. Also, they include negative rewards for incorrect actions and present a partially observable world through a grid-based observation of size $7\times7$. These environments can be adjusted in size and complexity, resulting in an ideal benchmark environment for curriculum optimization.

Throughout our work, we will be using the \textit{DoorKey} and \textit{DynamicObstacles} environments, which require the agent to learn to explore its environment to find the exit. In the \textit{DoorKey} environment, the agent needs to additionally search for a key to open doors blocking its way, while in \textit{DynamicObstacles} it needs to avoid randomly moving opponents. Both environment types are shown in~\Cref{fig:Different sizes of the Minigrid} including different level sizes. The observation space in these Minigrid environments consists of the direction of the agent, a $7\times 7\times3$ RGB representation, and the mission of the agent, providing details about objects, colors, and states within the agent's view. The action space consists of seven actions, which include moving forward, turning left or right, picking up or dropping objects underneath the agent, and two placeholders.

\todo{In the implementation the PPO in Leon's thesis does not use anything aprt from the PPO on CNN}

\section{Related Work}

In the following, we will provide an overview of related work in the domain of automated curriculum scheduling/learning. In the context of each work, we describe the differences to the proposed RHEA CL algorithm, which will be presented in the subsequent section.

Recent research has illustrated a diverse range of CL techniques, yet there is no consensus on a definitive universally "best" curriculum scheduling method. Surveys conducted by Wang et al. \cite{wang2021survey}, Soviany et al. \cite{soviany2022curriculum}, and Narvekar et al. \cite{narvekar2020curriculum} all highlight this fact and draw attention to the vast variety of CL approaches. 
Direct comparisons between these methods are largely absent, primarily because of their niche applications. This scenario emphasizes the necessity of creating flexible CL strategies, such as RHEA CL. Unlike traditional methods which require manual curriculum and precise difficulty-based adjustments, RHEA CL automates this process using EAs eliminating the need for difficulty assessment. 
Teacher-Student Curriculum Learning (TSCL) \cite{matiisen2019teacher} and Self-Paced Curriculum Learning (SPCL) \cite{jiang2015self} methods, rely on more structured linear progression models, in contrast to the automated RHEA CL. TSCL and SPCL are very effective in specific contexts, but these methods require significant upfront knowledge and may not adapt well to evolving learning scenarios.

Unlike the adversarially trained generator network method by Florensa et al. \cite{florensa2018automatic} that autonomously discovers and learns a diverse set of tasks, RHEA CL uses evolutionary algorithms with curriculum learning to adaptively optimize task difficulty and enhance agent performance in complex environments. Eimer et al. \cite{eimer2021self} introduce a framework for automatic curriculum learning that focuses on optimizing the sequence of tasks with the uses of self-paced learning in contrast to RHEA. Parker-Holder et al. \cite{parker2022evolving} automatic curriculum learning using reinforcement learning to improve the curriculum. RHEA CL's flexibility and adaptability fill a crucial gap in the literature, offering a novel solution to the inefficiencies associated with traditional CL methods and promising more effective learning across multiple domains.

\section{Methodology}

The proposed RHEA CL algorithm incorporates a task-driven curriculum learning approach, optimized for environments with variable complexities, such as the ones provided in the Minigrid framework utilized in this study. This method intelligently adapts to increasing task difficulties as the environmental size expands, necessitating a flexible and scalable learning strategy.

At its core RHEA CL is an EA that works in conjunction with a curriculum scheduler to systematically enhance the long-term performance gains of the trained RL agent. Throughout this work, PPO will be used as the RL agent to be trained. Note that RHEA CL is not limited to this choice and may be combined with any RL algorithm. PPO ensures a balanced exploration of new actions and maximization of cumulative rewards while maintaining policy updates within a conservative range to prevent detrimental performance leaps. 

The optimization process starts with the initialization of a population of curricula at the onset of each epoch, followed by a rewards matrix reset to track performance across generations. A curriculum is represented as a list of levels to train on one after another during the next training iterations. Our goal in the selection of levels is to maximize rewards across all environments. Although our primary focus is on single-objective optimization, multi-objective optimization using NSGA-II~\cite{deb2000fast} has been explored to assess its impact on the algorithm's performance. Therefore, Genetic Algorithms (GA, single-objective) and the Non-dominated Sorting Genetic Algorithm-II (NSGA-II, multi-objective) have been tested for the search for optimal curriculum configurations.

To ensure that the first step of the horizon is valued the most, we calculate the discounted return of a curriculum for which the rewards of each curriculum step are weighted with a dampening factor $(\gamma)$ that is exponentially decreasing over time. Similar to the discount factor in reinforcement learning, the dampening factor places greater emphasis on immediate steps over future ones. The curriculum score of curriculum $i$ is given by: 
\begin{equation}
    \textit{curriculumScore}_i = \sum_{j=0}^{\text{curricLength}} \textit{reward}_j \cdot \gamma^j.
\end{equation}
In the single-objective approach, $\textit{reward}_j$ describes the average performance in all environments after training in the j-th curriculum segment.
In the multi-objective scenario, we returned one $curriculumScore$ for each environment, instead.
Early tests have shown that for the small number of environments used in our tests, NSGA-II has resulted in no noticeable gains and will therefore not be considered throughout the remainder of this work.

Our approach needs to carefully balance many hyperparameters including the curriculum's length, the number of generations per epoch, and the number of training steps within each curriculum segment. This balance ensures comprehensive training across all environments, emphasizing the agent's adaptability and proficiency in varied tasks. Furthermore, we may allow the curriculum to include multiple environments per training step. This is especially relevant when the agent has to transition from a simple level to a more complex one. Training on both in quick succession directly targets the agent's adaptability ensuring that previously learned skills are not forgotten and the learning curve is not too steep. Therefore, we introduce a parameter \textit{(paraEnv)} for setting a maximal number of environments to train on in parallel.
The final key parameters for many evolutionary optimization approaches are the crossover and mutation rates, which dictate the evolutionary dynamics, influencing the combination and evolution of curricula across generations.
\Cref{sec:hyperparameter-optimization} describes our structured experiments on tuning the algorithm's parameters and getting a better understanding of their impact on its performance.

The method can be summarized using the pseudo-code shown in Algorithm \ref{alg:rhea_cl}. The variable \(epoch\) represents each cycle of the algorithm's operation, where a series of generations are processed to evolve the population of curricula. \(Population\) refers to the set of curricula being optimized. The \(Rewards Matrix\) is a matrix to record the performance of each curriculum across generations, initialized to zero at the start of each epoch. Within the context of generations (\(gen\)), each \(Curriculum\) obtained from the evolutionary algorithm undergoes evaluation, where its effectiveness is measured and accumulated in the Rewards Matrix. \(curricLength\) denotes the number of steps or tasks within a single curriculum, guiding the training process. The \(reward_j\) captures the performance score of the \(j^{th}\) step in a curriculum, adjusted by a dampening factor (\(\gamma_j\)) to prioritize immediate gains. \(nextStartingPoint\) is determined by selecting the curriculum with the highest cumulative reward, indicating the most effective training sequence for subsequent epochs. The \(Rewards Matrix\) is of the shape \(nGen\) x \(curricCount\) for single objective, and \(nGen\) x \(curricCount\) x \(numObj\) for multi-objective, where \(nGen\) is the number of generations in an epoch, \(curricCount\) is the number of curricula used, and \(numObj\) denotes the number of objectives used. An important step in the pseudo-code at the end of each epoch is determining the next starting point with a method \(GetBestCurriculum\) that takes the rewards matrix as a parameter. \(GetBestCurriculum\) outputs the curriculum with the highest rewards and this curriculum is selected as the starting point of the next epoch. 
The iterative refinement of curricula, underscored by the algorithm's strategic evaluation and selection mechanisms, showcases a systematic approach to enhancing reinforcement learning models' performance through curriculum learning and evolutionary optimization.

\begin{algorithm}[t]
\caption{RHEA CL General Procedure}
\label{alg:rhea_cl}
\footnotesize
\begin{algorithmic}
\FOR{\textbf{each} epoch}
\STATE Initialize Population
\STATE Initialize Rewards Matrix rewards[][] with 0’s
\FOR{\textbf{each} gen \textbf{in} Generations}
\FOR{\textbf{each} Curriculum \textbf{in} Population}
\FOR{\(j = 0\) \TO curricLength}
\STATE Train on the Curriculum step\STATE \(reward_j = \text{Evaluate Curriculum Step } j\)
\STATE \(rewards[gen][j] \mathrel{+}= reward_j \cdot \gamma^j\)
\ENDFOR
\ENDFOR
\STATE Select the best Individuals for \textit{Reproduction}
\STATE Apply \textit{Crossover} and \textit{Mutation}
\STATE Update Population
\ENDFOR
\STATE Train on first environment of best known curriculum
\ENDFOR
\end{algorithmic}
\end{algorithm}

\todo[]{Where's the fig 2 coming from?}

\section{Experiments and Results}
\label{sec:experiments}
\todo[inline]{I started to describe the results. You can easily spot sections I wrote, by their comment before the start of the paragraph. I like to use those to give a one-line summary of what I wrote. Experiment Setup and Hyperparameter Optimization are mostly done}

To test the proposed RHEA CL algorithm, we selected two environments of the Minigrid framework, namely, \textit{DoorKey} and \textit{DynamicObstacles}. 
In both environments, we control the task's difficulty by changing the size of the room layout and the agent's vision range. To reduce the run-time of our experiments while still being able to study levels of increasing difficulty, we chose to reduce the default vision range from a $7 \time 7$ grid to a $5 \times 5$ grid. At the same time, we used levels of size $6\times6, 8\times8, 10\times10, 12\times12$. 
Since the agent's view is based on its current position and direction, it is only able to partially observe a given state. Therefore, When increasing a level's size, it is harder to find points of interest such as the key or the exit, and avoid obstacles that are outside the agent's vision. While those levels are comparatively small, they allow for quick iterations and a more fine-grained analysis of the curricula optimization.

For PPO~\cite{schulman2017proximal} experiments in our evaluation we use a set of default parameters provided by developers of Minigrid~\cite{Willems_2018}. \Cref{tab:parameters-ppo} shows these recommended default parameters. Given some initial experiments, these parameters indeed appeared to work decently for all our environments and level sizes even if not solving them entirely in all cases. This leaves us with enough room to study the impact of the curriculum and its online optimization.

\begin{table}[t]
    \centering
    \caption{PPO Hyperparameters}
    \label{tab:parameters-ppo}
    \begin{tabular}{c|c}
         \toprule
         Parameter & Default Value \\
         \midrule
         Batch Size & 256 \\
         Discount Factor & 0.99 \\
         Learning Rate & 0.001 \\
         Generalized Advantage Estimator & 0.95 \\
         Entropy Coefficient & 0.01 \\
         Value Loss Coefficient & 0.5 \\
         Maximum Gradient Norm & 0.5 \\
         Adam Stability Factor $\varepsilon$ & 1e-8 \\ 
         Adam Smoothing Factor $\alpha$ & 0.99 \\ 
         Clip $\varepsilon$ & 0.2 \\
         Frames per Process & 128 \\
         \bottomrule
    \end{tabular}
    \todo{check for correct var}
\end{table}

\begin{figure*}
    \centering
    
    \begin{subfigure}[b]{0.32\textwidth}
         \centering
         \includegraphics[width=1.0\textwidth]{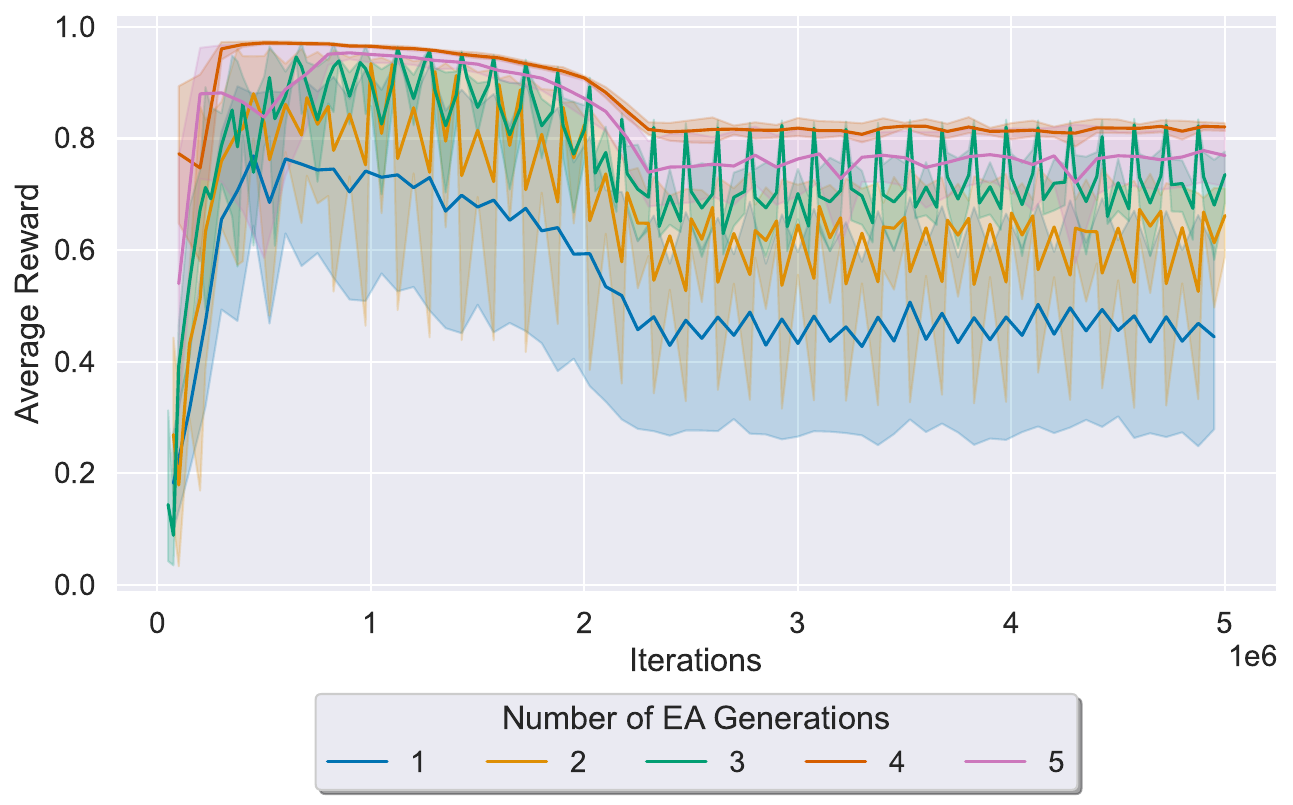}
         \caption{Number of EA Generations}
         \label{fig:hyperparameters_number_of_generations}
     \end{subfigure}
     \hfill
     \begin{subfigure}[b]{0.32\textwidth}
         \centering
         \includegraphics[width=1.0\textwidth]{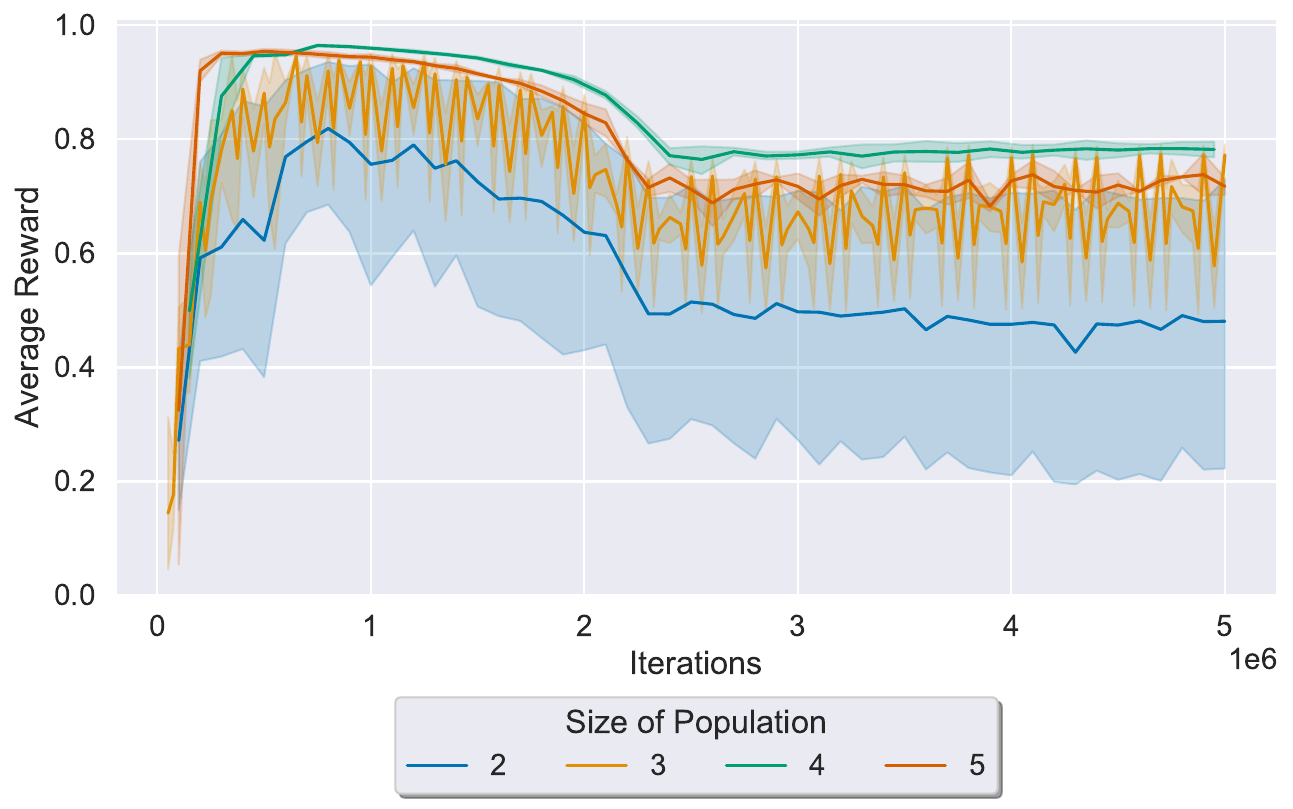}
         \caption{Size of Population}
         \label{fig:hyperparameters_size_of_population}
     \end{subfigure}
     \hfill
     \begin{subfigure}[b]{0.32\textwidth}
         \centering
         \includegraphics[width=1.0\textwidth]{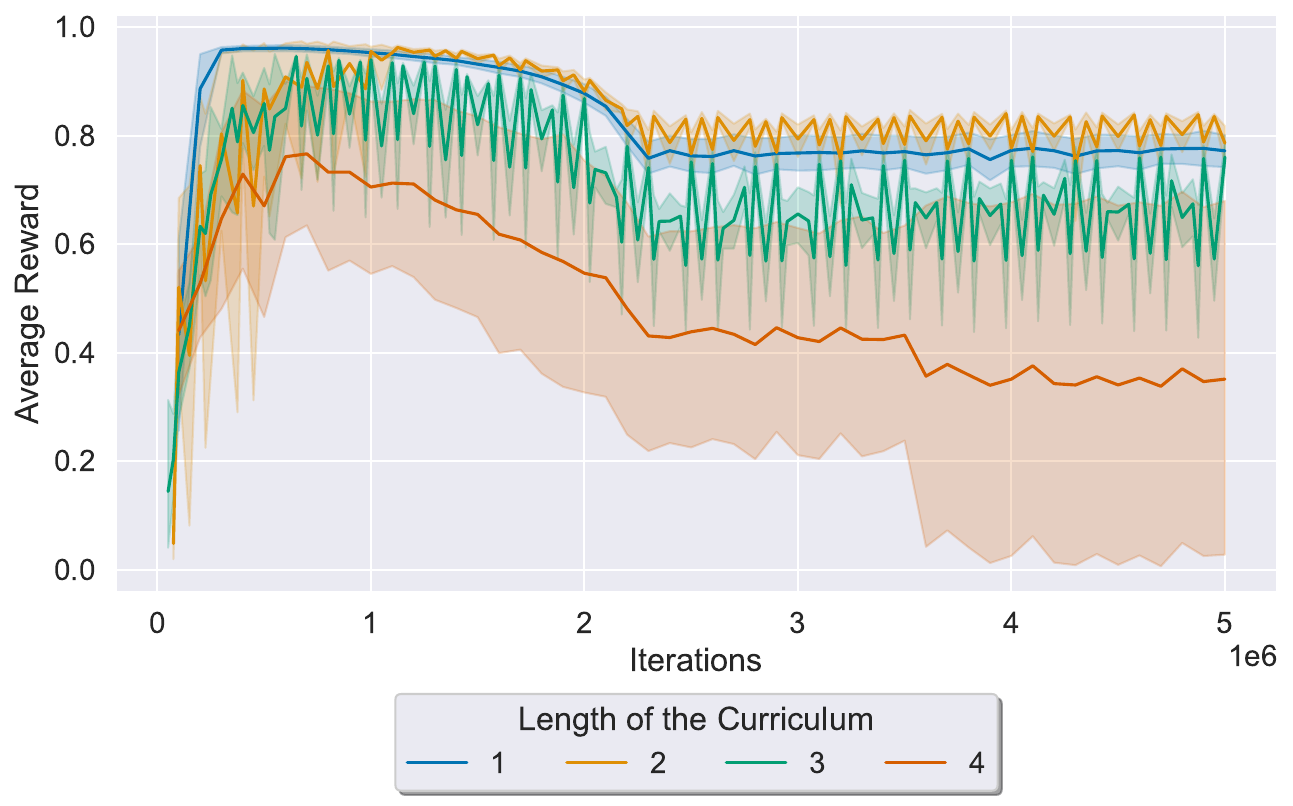}
         \caption{Length of Curriculum}
         \label{fig:hyperparameters_length_of_curriculum}
     \end{subfigure}

     \begin{subfigure}[t]{0.32\textwidth}
         \centering
         \includegraphics[width=1.0\textwidth,clip,trim=0cm -2cm 0cm 0cm]{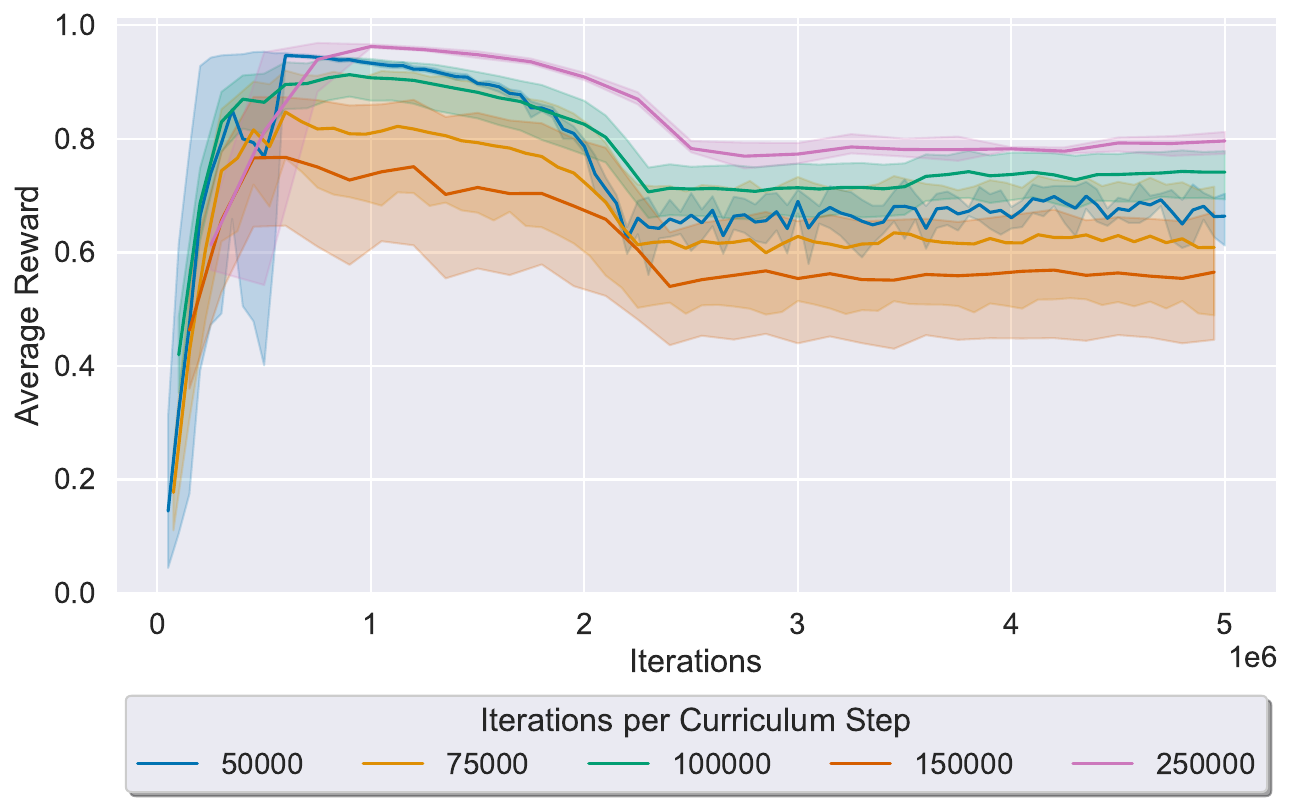}
         \caption{Iterations per Curriculum Step}
         \label{fig:hyperparameters_curriculum_steps}
     \end{subfigure}
     \hspace{0.1\textwidth}
     \begin{subfigure}[t]{0.34\textwidth}
         \centering
         \includegraphics[width=1.0\textwidth,clip,trim=0cm 0cm 0cm 0cm]{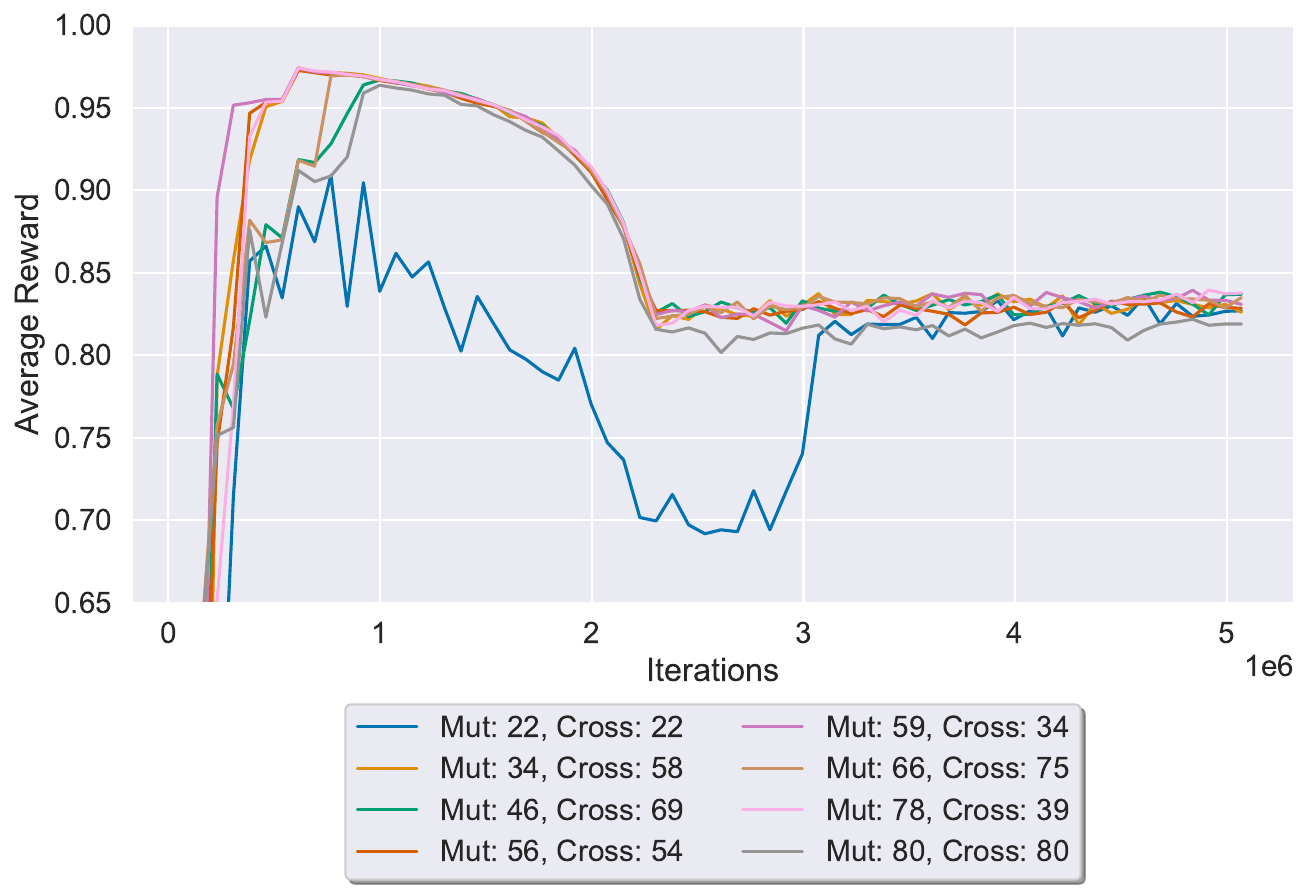}
         \caption{Mutation and Crossover rates}
         \label{fig:hyperparameters_mutation_and_crossover_rates}
     \end{subfigure}
    \caption{Results of the hyperparameter optimization.}
    \label{fig:hyperparameters}    
\end{figure*}

\subsection{Hyperparameter Optimization}
\label{sec:hyperparameter-optimization}

Before comparing our proposed algorithm to other curriculum optimizers, we analyzed the impact of its hyperparameters. Due to resource constraints, a comprehensive exploration of the entire hyperparameter space was not feasible. Consequently, the optimization process was conducted with a limited subset of potential parameter values. As part of this analysis, we trained agents with the parameter combinations that are listed in the appendix (see \Cref{tab:rhea_cl_ga_doorkey}). For each combination, we aggregated the performance across 5 random seeds. The following plots further aggregate the results by the parameter in question. Due to the aggregation of agents with different numbers of iterations per evaluation, the lines may look jaggy. However, their convergence tells a clear picture.

\paragraph{Number of Generations $(nGen)$}

The quantity of EA generations significantly influences the time expended in discovering the optimal curriculum for any given time step. Intuitively, a higher number of generations is expected to enhance performance.

To scrutinize this hypothesis, we examined values within the range $[1,5]\in\mathbb{N}$. The outcomes, depicted in \Cref{fig:hyperparameters_number_of_generations}, broadly align with the anticipated behavior. Nonetheless, the efficacy of employing additional generations appears to plateau. Notably, employing four generations yielded marginally superior results compared to five generations, although this discrepancy might stem from sampling variability.

When deciding on the number of generations, it is necessary to keep in mind the heavy burden on computation time higher values will cause. We suggest keeping an eye on performance improvements during the early iterations and deciding based on the results which setting to go with.

\paragraph{Size of Population $(curricCount)$}

Similar to the number of generations, the size of the population in evolutionary algorithms can significantly impact the optimization process. We conducted experiments varying the population size within the range $[2,5]\in\mathbb{N}$. 

A comparison of aggregated results per population size is presented in \Cref{fig:hyperparameters_size_of_population}. Despite large fluctuations for a population size of 2, the overall performance remained relatively stable across tested population sizes. The former can be caused by bad initialization of the population and should be effectively avoided by using a sufficiently large population size.

\paragraph{Length of Curriculum $(curricLength)$}

The length of the curriculum describes the horizon covered by a single curriculum. We test values in the range $\lbrack 1, 5\rbrack \in \mathbb{N}$. Setting the curriculum length too high results in excessively long horizons and may lead the agent into local optima. Furthermore, it results in significant computational costs due to the need for more training iterations and evaluations. In contrast, a low curriculum length diminishes the return of curriculum optimization.

\Cref{fig:hyperparameters_length_of_curriculum} shows the results aggregated by curriculum length. We were surprised to see, that the performance decreases so drastically for a curriculum length of 3 and 4. Nevertheless, a length of 2 has shown slight improvements in performance in comparison to a greedy selection caused by a curriculum length of 1. Further, improvements may be possible by changing how mutations and crossover are done. However, this will be a line of future work.

\paragraph{Iterations per Curriculum Step $(iterSteps)$}

Similar to the length of the curriculum, the number of iterations per curriculum step impacts the length of the horizon. We compared the values $\lbrace 25k, 50k, 75k, 100k, 150k, 250k\rbrace$. Given an overall training duration of $5m$ iterations, choosing higher values per step, results in fewer runs of the curriculum optimizer, resulting in a narrower distribution in the graphs. This occurs because each selected curriculum remains active for an extended period.

The results shown in \Cref{fig:hyperparameters_curriculum_steps} do not show a clear answer for which setting is best. With an increasing number of iterations per curriculum step, the results seem to cause a larger spread in performance. Conversely, lower values provide increased opportunities for course corrections during training, especially in cases of unfavorable epochs or when encountering bad initialization of the EA.

\paragraph{Mutation and Crossover}

Also, we evaluated different rates for mutation and crossover. Parameter combinations have been chosen using Sobol sampling \cite{sobol1967distribution} to ensure that the parameter space is evenly covered. 
To increase the comparability of results, we kept the other parameters fixed at number of generations = 3, size of population = 3, length of curriculum = 3, and iterations per curriculum step = 75k.
\Cref{fig:hyperparameters_mutation_and_crossover_rates} shows the aggregated results.

It appears to be the case that setting both rates to medium rates, on average achieves the best performance. E.g. the run with a mutation rate of $56\%$ and crossover rate of $0.54\%$ as well as the run with $59\%$ and $34\%$ achieve the peak in performance the fastest. 

For the interpretation of these results, we need to consider the rather low number of generations used in curriculum optimization. Consequently, the effects of crossover and mutation might not be as pronounced within this time frame. Similarly, tuning the mutation and crossover rates becomes more important for higher numbers of generations. Further optimizations may be achievable by adjusting the type of mutation and crossover~\cite{Doc2022a}.



\todo{The graph showing the variation of gamma can be added as there's some space left}

\paragraph{Parallel Training} Although not included in this hyperparameter evaluation, we want to report the benefits of training on multiple levels in parallel. Due to the low number of environments, we maintained a consistent value of 2 levels per curriculum step. While this parameter wasn't directly evaluated, its implementation proved beneficial by expanding the spectrum of possible curricula and thereby increasing the search space. Combined with the benefits of emphasizing smoother transitions between difficulty levels, this contributed to achieving a more stable performance overall.


\subsection{Comparison to other Curriculum Optimizer}

We compare our method to several baselines, i.e.:
\begin{itemize}
    \item \textbf{Rolling Horizon Random Search (RHRS):} To test the impact of the EA optimization, we are using a similar process to RHEA CL but replaced the EA component with a random curriculum sampling.
        
    \item \textbf{All Parallel (AllParallel):} Further, we test if the dynamic selection of levels to train on benefits the agent's training in comparison to training on all four in quick succession. Given the set of all level sizes, we iterate between levels after each episode. Once all levels have been trained, the process is repeated.

    \item \textbf{Self-Paced Curriculum (SPCL) \cite{jiang2015self}:} We also compare RHEA CL to a standard curriculum learning approach that adapts the environment based on the agent's current performance. Starting with the smallest level size, the level size is increased once the agent's performance exceeds $85\%$ of the maximum possible reward and decreases once it falls below $50\%$. Updates of the level size are evaluated after 25k iterations, potentially giving the agent a chance to quickly adjust to its current capabilities.

    \item \textbf{PPO \cite{schulman2017proximal}:} Finally we compare RHEA CL to a vanilla PPO agent (No Curriculum) with a configuration that uses the default hyperparameters recommended by the authors for the said Minigrid environments \cite{Willems_2018}. The agent is trained in an environment of 12x12 grid size. 
\end{itemize}

During the evaluation, each configuration was tested 5 times each with different seeds. In each run, we step-wise decrease the maximum number of steps until the task is considered to be failed. This aims at increasing the pressure to perform well across all level sizes.

\subsection{Performance Analysis}

\begin{figure*}
    \begin{subfigure}[b]{0.45\textwidth}
         \centering
         \includegraphics[width=1.0\textwidth]{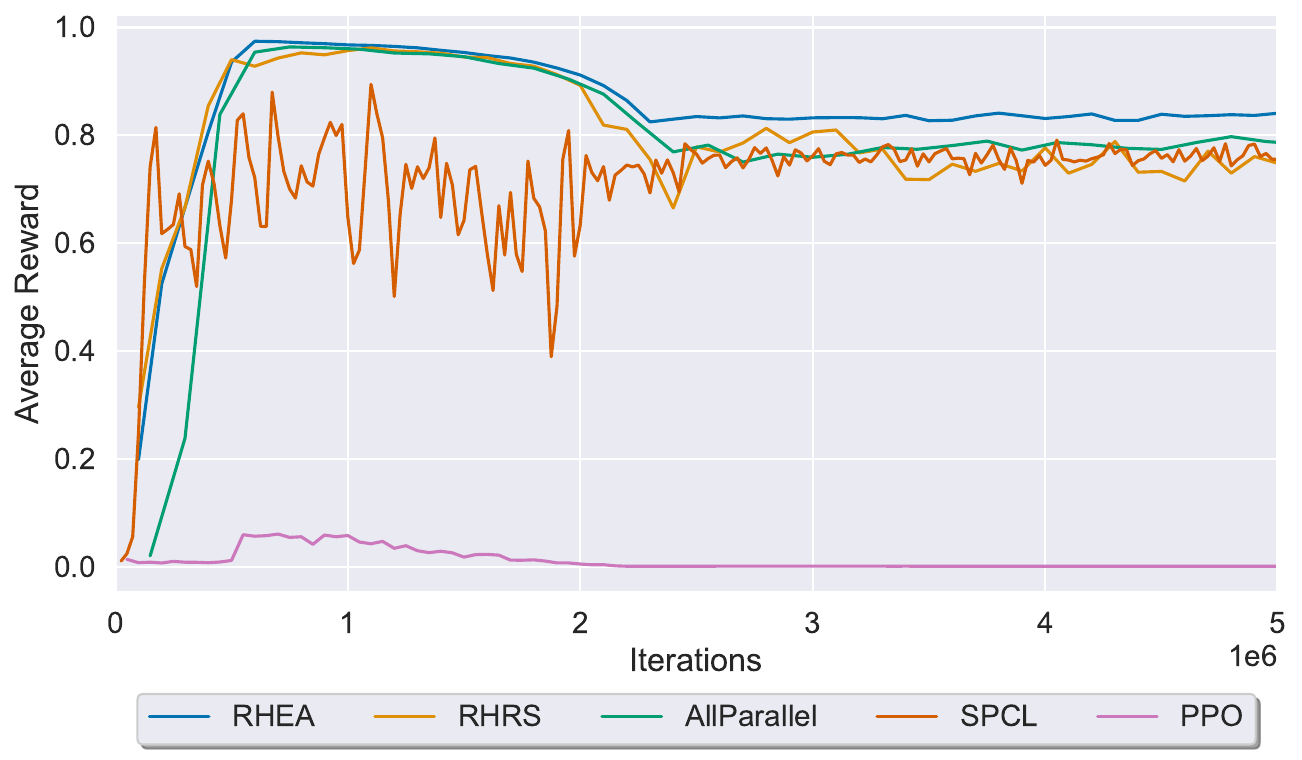}
         \caption{Results DoorKey}
         \label{fig:res-door-key}
     \end{subfigure}
     \hfill
     \begin{subfigure}[b]{0.45\textwidth}
         \centering
         \includegraphics[width=1.0\textwidth]{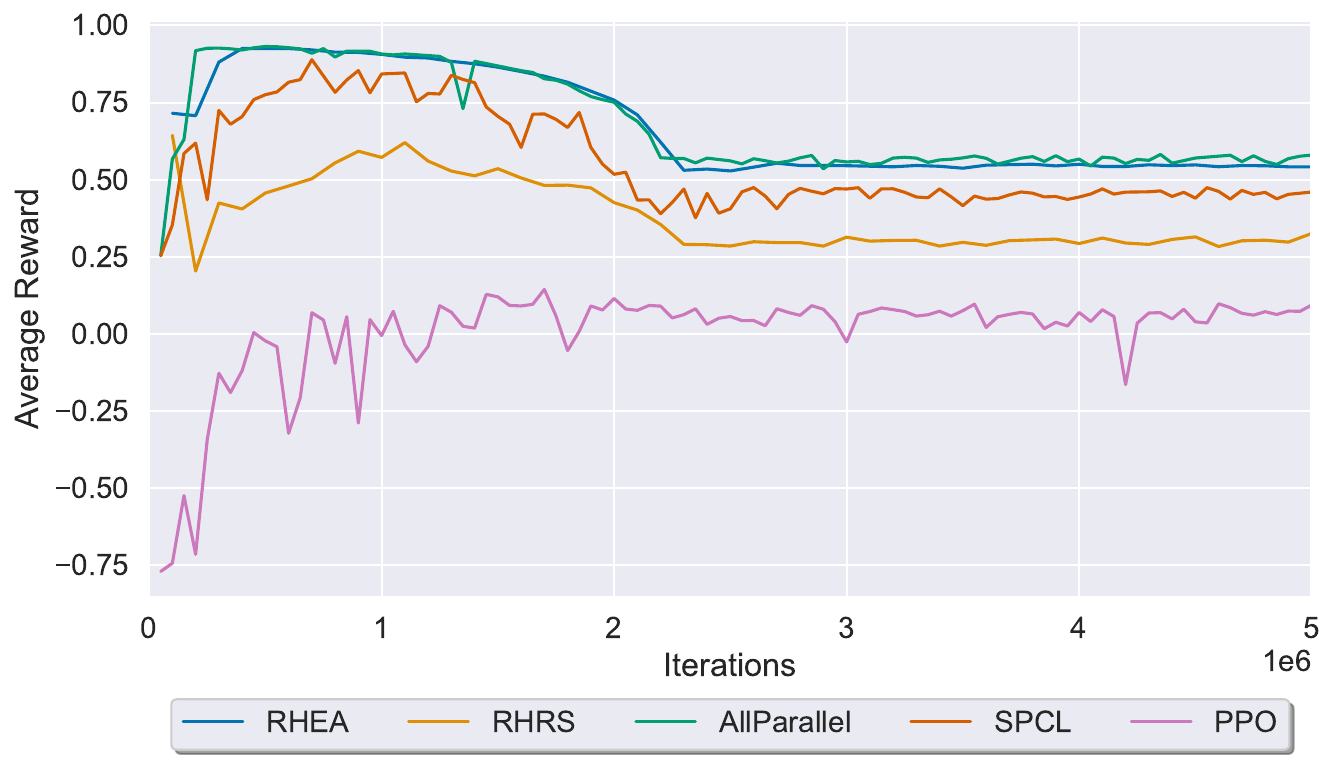}
         \caption{Results Dynamic Obstacles}
         \label{fig:res-dynamic-obstacles}
     \end{subfigure}
     \todo[inline]{RHEA color in the left plot is still wrong}
     \caption{Comparing test performance of tested curriculum learning methods.}
     \label{fig:res-performance}
\end{figure*}

\Cref{fig:res-performance} shows the agent's performance across all difficulty levels of an environment throughout training. The curves indicate different curriculum styles and the performance of the vanilla PPO agent. The proximity of the curves to each other complicates the interpretation of the graphs, especially when the standard deviation is included. \Cref{fig:res-performance} shows the curves without the standard deviation with 5 different seeds. \Cref{fig:res-performance_with_std} shows plots with standard deviation.
Most of the following plots display similar behavior, where they go up at first to nearly $\text{100\% }$ performance, and then after roughly 1.1 million iterations, they start to take a dip in their performance until it later stabilizes again. Although looking strange at first glance, this happens due to the decreasing number of maximum steps allowed per environment. Even though the performance looks slightly worse, the agent is getting better at solving the task, because wasting time is punished more and more, which seems desirable to have either way. The \textit{maxSteps} remain at the default values until the \textit{iterationsDone (iD)} reach 500,000. After this, it starts reducing and is given by: 
\begin{equation}
    \textit{maxSteps}_{\textit{new}} = \textit{maxSteps}_{\text{default}} \cdot \max\left(1 - \frac{\textit{iD} - 5 \cdot 10^5}{2 \cdot 10^6}, 0.15\right).
\end{equation} This is done so that the agent first learns the
basic tasks before it gets more difficult with the reduced \textit{maxSteps}.
 Furthermore, a lower score with lower maximum steps allowed is to be expected, as the reward is calculated as follows: $reward =  1 - 0.9 \times (takenSteps / maxSteps)$, so by having lower $maxSteps$, the agent just gets penalized faster and is not necessarily getting worse. 

The comparison between RHEA CL and its baselines, RHRS, SPCL, and All Parallel, reveals some subtle and some very stark performance differences across various configurations and environments. RHRS is closely related to RHEA CL but exhibits instability and inconsistent performance and underperforms as compared to RHEA CL. RHEA CL demonstrated smoother, more stable outcomes, particularly in the DoorKey and DynamicObstacles environments, suggesting the benefits of using an EA for the optimization of its curricula. SPCL, which adjusts training environments based on agent performance, showed erratic behavior and lower peaks than RHEA CL, particularly in the DoorKey environment, and was generally less stable. The All Parallel approach, utilizing all environments simultaneously during training, offered a more stable and less erratic performance than SPCL, marginally outperforming RHEA CL in the DynamicObstacles environment but falling short in the DoorKey scenario, highlighting the effectiveness of RHEA CL's methodological advantages in achieving superior and more consistent results. The vanilla PPO which was trained in the largest environment of both tasks failed to achieve any meaningful performance in the given number of iterations, highlighting the importance of curriculum learning.

\todo{add the function to show the reduction in the max number of steps allowed for the EAs and remove the figures from the text}

\section{Limitations and Future Work}

The RHEA CL approach, while outperforming baselines like PPO and SPCL in certain environments, still has room for growth. Future work could examine how different neural network architectures or hyperparameter tuning might improve performance, especially in complex scenarios such as DynamicObstacles. Additionally, adapting RHEA CL for various tasks and examining hyperparameter effects will become essential for enhancing its ease of use, as well as its versatility and robustness.

Notably, RHEA CL required significantly longer training times and it reflects a trade-off between comprehensive training and efficiency. This characteristic may benefit non-expert users by providing a more accessible entry point into sophisticated computational methods, like automated machine learning's role in machine learning. However, future enhancements should aim to streamline RHEA CL, potentially through more efficient algorithms, pre-trained models, or user feedback integration, without sacrificing its user-friendliness.


Improvements could include optimizing training environment access, adjusting challenge levels throughout training, and tailoring iterations to training stages. Innovating the evolutionary algorithm's fitness function and curriculum sampling to consider performance history could further refine curriculum optimization. A combined strategy of using RHEA CL in initial training phases before switching to more resource-efficient methods may also be advantageous. Broadening experimental domains will help confirm RHEA CL's adaptability and deepen insights into its learning optimization capabilities.

\section{Conclusion}

In this work, we proposed RHEA CL for the optimization of an agent's curriculum during training time. A careful analysis of the method's hyperparameters has shown a notable impact on the performance of RHEA CL and resulted in rough guidelines for their setup. Our tests have shown that some combinations exhibit faster convergence and better performance, but usually come with increased computational costs. Overall, RHEA CL seems to have a slight edge over other tested curriculum learners when it comes to early iterations of the agent's training and its final performance. Depending on the use case, the latter may justify increased computational costs to achieve performance gains that otherwise would have been unachievable due to the avoidance of local optima during RHEA CL's training process. 
In the future, we plan to analyze how knowledge of the hyperparameter landscape~\cite{MohBen2023} can be exploited in the optimization process.


\bibliographystyle{ieeetr}

\bibliography{references}

\appendix

\subsection{Comparison of performance with baselines (including standard deviations)}
\vspace{-3pt}

\begin{figure}[htbp]
     \centering
     \begin{subfigure}[c]{0.45\textwidth}
         \centering
         \includegraphics[width=1\textwidth]{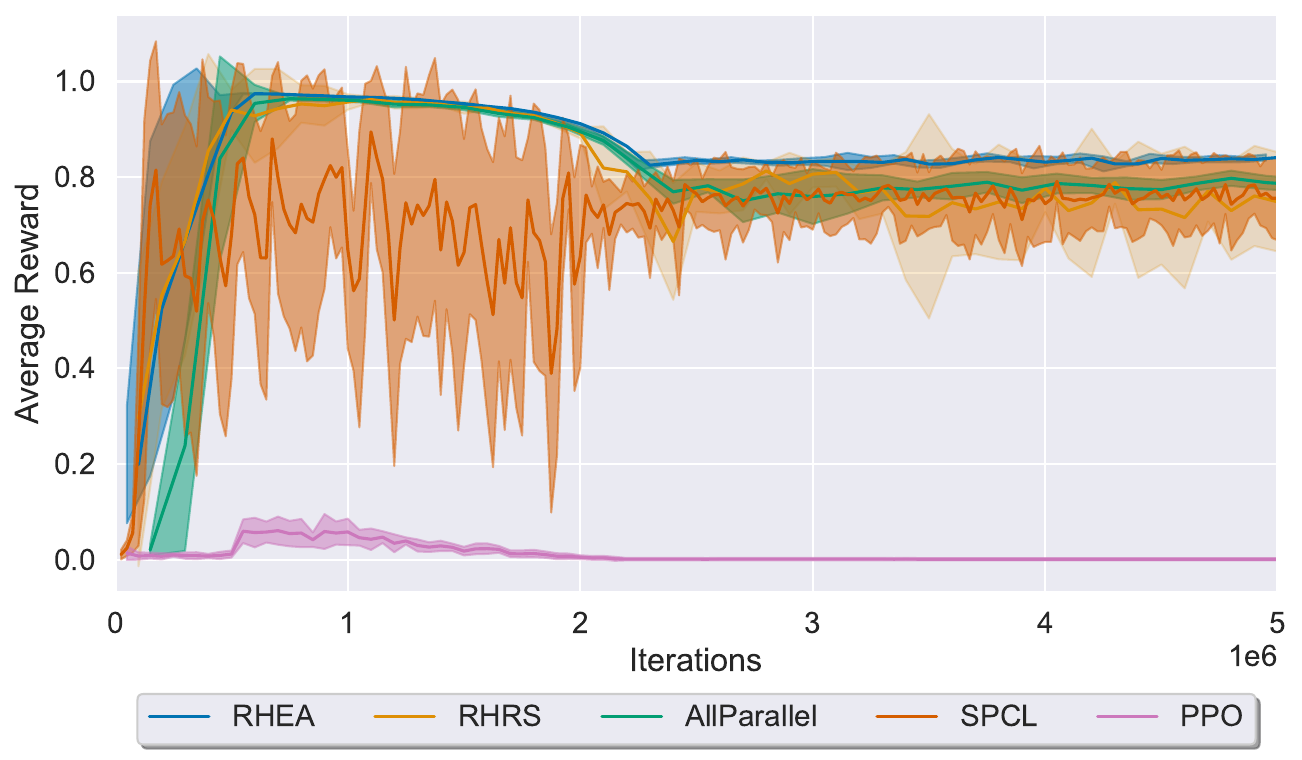}
         \caption{Results DoorKey}
         \label{figres-perf-DK-std}
     \end{subfigure}
     \hfill
          \begin{subfigure}[c]{0.45\textwidth}
         \centering
         \includegraphics[width=1\textwidth]{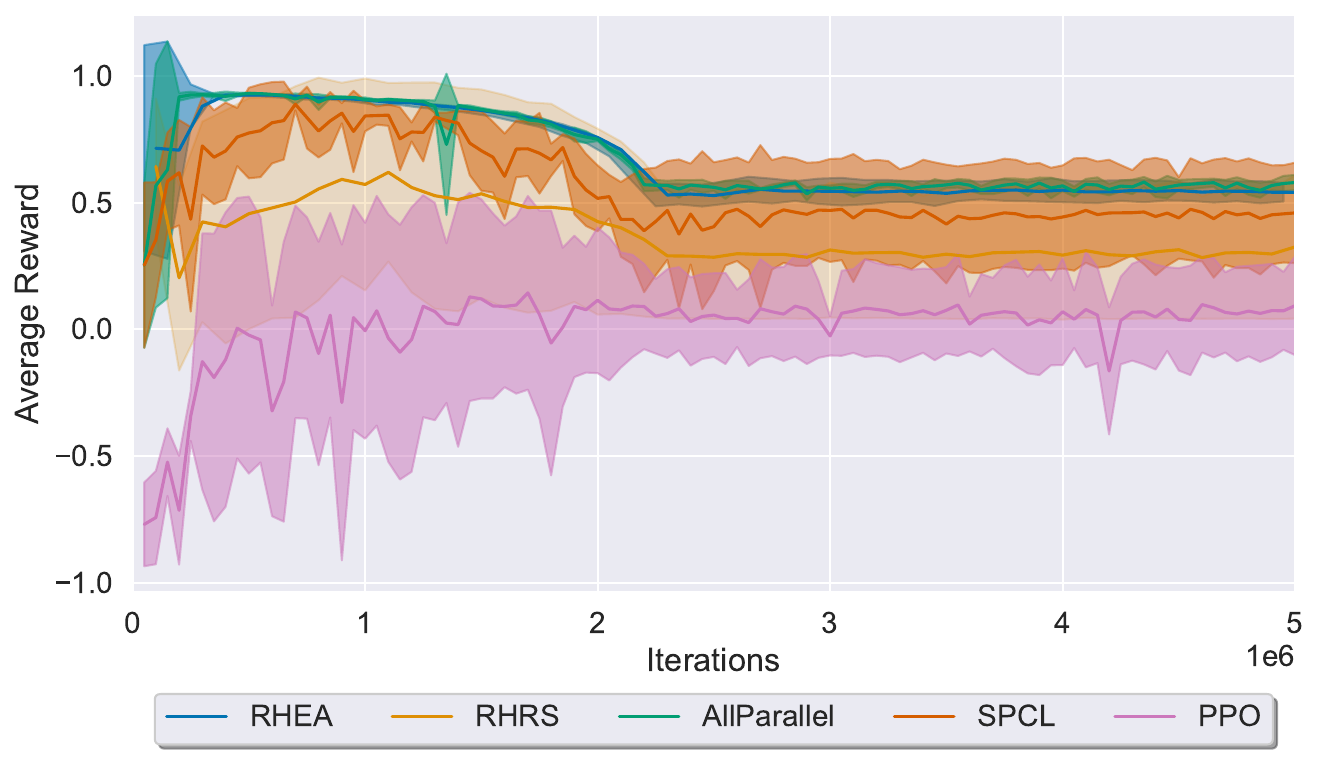}
         \caption{Results DynamicObstacles}
         \label{fig:res-perf-DO-std}
     \end{subfigure}
        \caption{Comparing test performance of tested curriculum learning methods with standard deviation.}
        \label{fig:res-performance_with_std}
\end{figure}

\Cref{fig:res-performance_with_std} shows the agent’s performance across all difficulty levels of an environment throughout training with standard deviation across 5 different seeds.
\vspace{-3pt}

\vspace{-20pt}

\subsection{Network Architecture}
The neural network used by the PPO agent consists of the following layers:
\begin{itemize}
    \item CNN with 3 input channels, 16 output channels, kernel size of 2, stride of 1, and no padding. The 3 correspond to the 3 channels in the observation space (5x5x3)
    ReLU
    \item MaxPooling layer with a 2x2 pool size
    \item CNN with 16 input channels and 64 output channels. Kernel stride and Padding values like above.
    \item ReLU
    \item Actor: 1 Linear Layer (64 in, 7 out) with tanh activation, 
    \item critic: 1 Linear Layer (64 in, 1 out) with tanh activation.
\end{itemize}
\vspace{-3pt}

\Cref{fig:network-architecture} shows a depiction of the resulting network architecture.
\vspace{-3pt}

\usetikzlibrary{positioning, shapes.geometric}

\begin{figure}[H]
    \centering
    \begin{tikzpicture}[node distance=0.2cm, scale=0.5, every node/.style={scale=0.7}]
        \node (input) [rectangle, draw] {Input 5x5x3};
        \node (conv1) [rectangle, draw, below=of input] {Conv1: 16 output, 2x2, stride 1};
        \node (relu1) [rectangle, draw, below=of conv1] {ReLU};
        \node (maxpool) [rectangle, draw, below=of relu1] {MaxPooling: 2x2};
        \node (conv2) [rectangle, draw, below=of maxpool] {Conv2: 64 output, 2x2, stride 1};
        \node (relu2) [rectangle, draw, below=of conv2] {ReLU};
        \node (actor) [rectangle, draw, below left=0.3cm and -0.3cm of relu2] {Actor: Linear 64x7, tanh};
            \node (critic) [rectangle, draw, below right=1cm and -1cm of relu2] {Critic: Linear 64x1, tanh};
    
        \draw[->] (input) -- (conv1);
        \draw[->] (conv1) -- (relu1);
        \draw[->] (relu1) -- (maxpool);
        \draw[->] (maxpool) -- (conv2);
        \draw[->] (conv2) -- (relu2);
        \draw[->] (relu2) -- (actor);
        \draw[->] (relu2) -- (critic);
    \end{tikzpicture}
    \caption{Network Architecture PPO}
    \label{fig:network-architecture}
\end{figure}
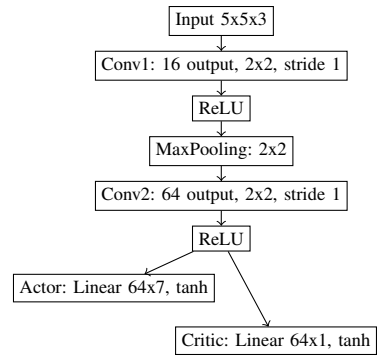
\vspace{-3pt}

\subsection{Tested RHEA CL Parameter Combinations}
\vspace{-3pt}
\begin{table}[H]
    \centering
    \caption{RHEA CL parameters tested for DoorKey}
    \label{tab:rhea_cl_ga_doorkey}
    \begin{tabular}{cccc}
    \toprule
    \textbf{Iterations per } & \textbf{Length of} & \textbf{Number of} & \textbf{Size of}  \\ 
\textbf{Curriculum Step} & \textbf{Curriculum} & \textbf{Generations} & \textbf{Population}\\
\midrule
    25 & 3 & 3 & 3 \\ \midrule
    50 & 3 & 3 & 3 \\ \midrule
    75 & 3 & 1 & 3 \\
    75 & 3 & 2 & 3 \\
    75 & 2 & 3 & 3 \\
    75 & 3 & 3 & 3 \\
    75 & 3 & 1 & 3 \\
    75 & 3 & 2 & 3 \\ \midrule
    100 & 3 & 3 & 3 \\
    100 & 4 & 3 & 2 \\
    100 & 1 & 3 & 3 \\
    100 & 1 & 3 & 5 \\
    100 & 2 & 5 & 2 \\
    100 & 3 & 2 & 3 \\
    100 & 3 & 3 & 3 \\
    100 & 4 & 3 & 3 \\
    100 & 1 & 3 & 3 \\
    100 & 1 & 5 & 3 \\
    100 & 2 & 3 & 3 \\
    100 & 2 & 4 & 3 \\
    100 & 3 & 3 & 3 \\ \midrule
    150 & 3 & 2 & 4 \\
    150 & 3 & 3 & 3 \\
    150 & 3 & 1 & 3 \\
    150 & 3 & 2 & 3 \\ \midrule
    250 & 3 & 3 & 3 \\ \bottomrule
    \end{tabular} 
\end{table}
\vspace{-3pt}

\begin{table}[h]
\centering
\caption{RHEA CL parameters tested for DynamicObstacles}
\label{tab:rhea_cl_dynamic_obstacle}
\begin{tabular}{cccc}
\toprule
\textbf{Iterations per } & \textbf{Length of} & \textbf{Number of} & \textbf{Size of} \\ 
\textbf{Curriculum Step} & \textbf{Curriculum} & \textbf{Generations} & \textbf{Population}\\
\midrule
50 & 3 & 3 & 3 \\
\midrule
75 & 3 & 3 & 3 \\
75 & 3 & 2 & 4 \\
75 & 3 & 2 & 3 \\
\midrule
100 & 3 & 2 & 4 \\
100 & 3 & 2 & 3 \\
100 & 3 & 3 & 3 \\
\midrule
150 & 3 & 3 & 3 \\
150 & 3 & 4 & 2 \\
150 & 2 & 4 & 3 \\
\bottomrule
\end{tabular}
\end{table}

\end{document}